\documentclass{article}

\usepackage{PRIMEarxiv}

\usepackage[utf8]{inputenc} 
\usepackage[T1]{fontenc}    
\usepackage{hyperref}       
\usepackage{url}            
\usepackage{booktabs}       
\usepackage{amsfonts}       
\usepackage{nicefrac}       
\usepackage{microtype}      
\usepackage{lipsum}
\usepackage{amsmath}
\usepackage{fancyhdr}       
\usepackage{graphicx}       
\graphicspath{{media/}}     

\title{Research on UAV Applications in Public Administration: Based on an Improved RRT Algorithm
\thanks{\textit{\underline{Citation}}: 
\textbf{Authors. Title. Pages.... DOI:000000/11111.}} 
}

\author{
  Zhanxi Xie \\
  Faculty of Humanities and Social Sciences  \\
  Macao Polytechnic University \\
  Macao, 999078 China\\
  \texttt{p2313081@mpu.edu.mo} \\
   \And
  Baili Lu \\
  College of Animal Science and Technology\\
  Zhongkai University of Agriculture and Engineering \\
  Guangzhou, 510225 China\\
  \texttt{18023304003@163.com} \\
   \And
  Yanzhao Gu \\
  Faculty of Applied Sciences \\
  Macao Polytechnic University \\
   Macao, 999078 China\\
  \texttt{p2311998@mpu.edu.mo} \\
   \And
  Zikun Li \\
  School of Economics and Management \\
  South China Normal University \\
  Guangzhou, 510006 China\\
  \texttt{18520610821@163.com} \\
  \And
  Junhao Wei\\
  Faculty of Applied Sciences \\
  Macao Polytechnic University \\
  Macao, 999078 China\\
  \texttt{p2312195@mpu.edu.mo} \\
  \And
  Ngai Cheong*\\
  Faculty of Applied Sciences \\
  Macao Polytechnic University \\
  Macao, 999078 China\\
  \texttt{ncheong@mpu.edu.mo} \\
}

\begin{document}
\maketitle

\begin{abstract}
This study investigates the application of unmanned aerial vehicles (UAVs) in public management, focusing on optimizing path planning to address challenges such as energy consumption, obstacle avoidance, and airspace constraints. As UAVs transition from 'technical tools' to 'governance infrastructure', driven by advancements in low-altitude economy policies and smart city demands, efficient path planning becomes critical. The research proposes an enhanced Rapidly-exploring Random Tree algorithm (dRRT), incorporating four strategies: Target Bias (to accelerate convergence), Dynamic Step Size (to balance exploration and obstacle navigation), Detour Priority (to prioritize horizontal detours over vertical ascents), and B-spline smoothing (to enhance path smoothness). Simulations in a 500 m³ urban environment with randomized buildings demonstrate dRRT's superiority over traditional RRT, A*, and Ant Colony Optimization (ACO). Results show dRRT achieves a 100\% success rate with an average runtime of 0.01468s, shorter path lengths, fewer waypoints, and smoother trajectories (maximum yaw angles <45°). Despite improvements, limitations include increased computational overhead from added mechanisms and potential local optima due to goal biasing. The study highlights dRRT's potential for efficient UAV deployment in public management scenarios like emergency response and traffic monitoring, while underscoring the need for integration with real-time obstacle avoidance frameworks. This work contributes to interdisciplinary advancements in urban governance, robotics, and computational optimization.

\end{abstract}


\section{Introduction}
\subsection{Research Background}
Unmanned Aerial Vehicle (UAV) as one of the most disruptive technologies since the early 21st century has spread widely in a short time from military field to civilian field and social service arena, driven by technical breaks in drone area introduced a notion of low-altitude economy opened up new solutions for the economic blockage. The National Comprehensive Tertiary Transportation Plan promoted the formation process 'low altitude economic' from provincial planning stage—national level in China on February, 2021; Until December, 2023 at the Central Economy Work Conference labeled 'Low Altitude Economics' as nationally strategic emerging industries, forming part of the 'new-quality productsivity', until Mar., 2024 integrating the notion in the governmental work report illustrated the importance of it with respect to our country's development.\par
The underlying trend of the Unarmed Aerial Vehicle is transitioning into 'governances infrastructure' from being merely an engineering application under impact due to rapidity of construction of smart city, frequent incidence of extreme weather event scenario's, refined needs for governance within society coupled technological innovation, societal need factors yet also posing its own sets of hurdles like technology gap challenges ethics controversies or institutional lagging situation.

\subsection{Research Significance}
\subsubsection{Theoretical Significance}
This research bridges critical gaps in technocratic governance theory by systematically examining UAV applications in public administration. Whereas traditional public administration theories prioritize institutional design and policy implementation, UAV technology drives smart governance through technological urbanization—transforming conventional urban spaces into intelligent systems. Furthermore, the integration of UAV technology with public management frameworks catalyzes cross-disciplinary innovation, fostering synergistic collaboration among administrative sciences, computer engineering, and legal studies.\par

\subsubsection{Practical significance}
UAVs can improve the public management level of disaster relief, dangerous environment operation and traffic surveillance at lower price and more reasonable resource allocation under certain conditions, going beyond the limitations.

\subsection{Research Problem}
Unmanned Aerial Vehicles (UAVs) demonstrate significant potential in aerial public management operations, particularly for emergency rescue missions and material delivery. However, widespread deployment faces three critical constraints:\par
\begin{itemize}
    \item \textbf{Energy limitations:} restricting operational endurance;
    \item \textbf{Urban obstacle:} density complicating navigation
    \item \textbf{Congested airspace:} requiring precise trajectory control
\end{itemize}

\section{Literature Review}
\subsection{Current Research on UAV Technology}
Unmanned Aerial Vehicle (UAV) path planning is the process of designing an optimal flight route for a UAV from a starting point to a target destination, taking into account constraints such as obstacle avoidance, energy consumption, and flight safety. Effective path planning allows UAVs to perform tasks with optimal efficiency, enhancing operational accuracy and reducing potential risks. By optimizing flight paths, UAVs can reduce travel distances, improve flight efficiency, and select the shortest, fastest, or most energy-efficient route, especially valuable in tasks such as delivery and inspection. This can minimize time wastage and energy expenditure. Path planning is critical to the broad application of UAV technology across various sectors. In logistics and delivery, agriculture, energy, and infrastructure monitoring, path planning allows for customized solutions that significantly boost operational efficiency. Additionally, UAV path planning plays a crucial role in environmental monitoring, disaster response, and smart city development.\par
Classical approaches to UAV path planning include optimization-based methods, graph-based methods, potential field approaches, and sampling-based methods. For instance, in recent years, metaheuristic algorithms have developed rapidly \cite{rwoa} \cite{lsewoa} \cite{lswoa} \cite{mrbmo} \cite{tswoa} \cite{gwoa}. Metaheuristic algorithms have been proven effective in UAV path planning \cite{ipso}. Wang et al. introduced a novel Matrix Aligned Dijkstra (MAD) algorithm, which employs high-dimensional matrix modeling for dynamic environments, coupled with GPU-accelerated forward exploration and backward navigation, achieving optimal UAV path planning under dynamic weather conditions \cite{bib12}. Li et al. proposed a 3D path planning model based on the R5DOS framework, combining an improved A* algorithm with the R5DOS cross model to reduce the number of search nodes, thereby significantly decreasing computational complexity and execution time \cite{bib13}. Jayaweera et al. developed a Dynamic Artificial Potential Field (D-APF) path planning method tailored for multi-rotor UAVs to track moving ground targets \cite{bib14}, while Li et al. proposed an improved Probabilistic Roadmap (IPRM) algorithm \cite{bib15} and the Feedback Rapidly-exploring Random Tree Star (FRRT*) algorithm \cite{bib17}. Furthermore, soft computing methods, including machine learning approaches like Q-learning \cite{qlearning} and DQN \cite{bib18} \cite{xgb}, as well as traditional convex optimization methods such as quadratic programming and linear programming, are commonly employed in UAV path planning. Hybrid approaches have also been proposed; for instance, combining RRT with the Artificial Potential Field (APF) method, Fan et al. \cite{bib20} developed a target-biased bidirectional APF-RRT* algorithm for UAV trajectory planning, incorporating a target bias strategy to increase random sample generation efficiency and using cubic spline interpolation to optimize the path, resulting in improved convergence speed and search performance. Another approach combines RRT with the Grey Wolf Optimizer (GWO); Kiani et al. \cite{bib21} presented three hybrid path planning methods based on an enhanced RRT algorithm coupled with variants of GWO (GWO, I-GWO, Ex-GWO), which enhance 3D autonomous robot path planning by optimizing path length and direction.\par


\subsection{UAV Applications in Public Administration}
Recent studies have demonstrated the wide-ranging potential of UAVs in various public management domains. In the field of land administration, UAV-based imagery and advanced image analysis techniques have been leveraged for cadastral mapping and boundary detection, enabling efficient and accurate spatial data collection \cite{bib1} \cite{bib2} \cite{bib3} \cite{bib4} \cite{bib5}. These studies explored methods such as contour extraction from UAV images, socio-technical assessments of UAV applications in land registration processes, and deep learning-based boundary delineation. For transportation-related applications, UAVs have been deployed for diverse purposes: structural displacement measurement of transport infrastructure \cite{bib6}, road safety monitoring and traffic flow analysis \cite{bib7}, as well as automated identification of hazardous obstructions at intersections using aerial data analytics \cite{bib8}. UAV imagery has also been utilized for detailed road surface condition assessment and pavement distress analysis through image-based 3D reconstruction techniques \cite{bib9}. Furthermore, coordinated multi-UAV systems have been investigated for traffic management and driver behavior detection, offering new opportunities for intelligent transportation systems \cite{bib10}. These advancements highlight UAVs’ capability to provide timely, high-resolution, and cost-effective data, thereby supporting decision-making processes in land governance, infrastructure maintenance, and transportation management.

\section{Methodology}
This study uses MATLAB R2023a to generate a 500 m³ 3D urban map with randomized building heights (18 m – 270 m). UAV start and end coordinates are set to (10, 10, 1) and (470, 420, 50), respectively. Traditional RRT and the proposed dRRT are simulated for path planning, with task completion time as the key metric.

\section{Simulation of Public Management Scenarios Using Enhanced RRT}
Path planning constitutes a critical enabling technology for ensuring efficient and secure UAV operations in urban environments. The presence of structural obstacles (e.g., buildings, bridges, power lines) necessitates collision-free trajectories while simultaneously requiring energy-optimal routing to extend mission endurance. Among existing solutions, the Rapidly-exploring Random Tree (RRT) algorithm has emerged as a prominent approach for UAV autonomous navigation due to its sampling efficiency in high-dimensional spaces. However, conventional RRT exhibits three fundamental limitations:\par
\begin{itemize}
    \item \textbf{Path quality:} The generated paths are not optimal and tend to be winding.
    \item \textbf{Lack of convergence:} RRT does not guarantee an optimal solution and may fail to find the shortest path in complex environments.
    \item \textbf{Non-deterministic quality:} RRT's randomness can result in unstable path quality and, with limited samples, may not find a solution.
\end{itemize}\par
In order to solve the problem that the general RRT Algorithm cannot overcome the shortcoming of the UAV Path Planning, the paper proposes a kind of enahnced RRT Algorithm (dRRT). The purpose is to overcome the defects of the original RRT Algorithm, which not only tries to solve the problems of obtaining feasible solutions faster and more practical than traditional RRT Algorithms, but also explores the practicability and feasibility of using the new type of algorithm-RRT-for search paths in UAV City flights.

\subsection{RRT Algorithm}
\begin{figure}[htbp]
    \centering
    \includegraphics[width=0.8\textwidth]{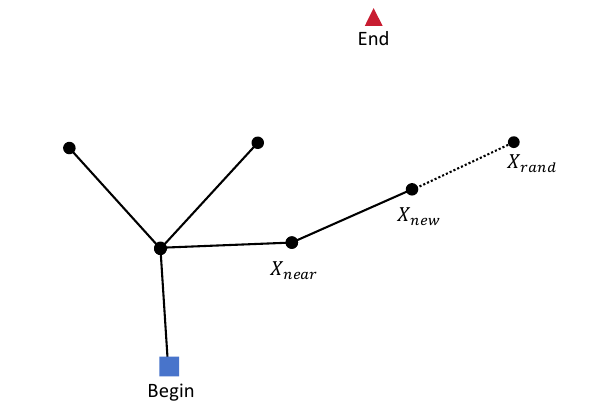}
    \caption{Extension process of RRT}
    \label{fig1}
\end{figure}

\begin{figure}[htbp]
    \centering
    \includegraphics[width=0.6\textwidth]{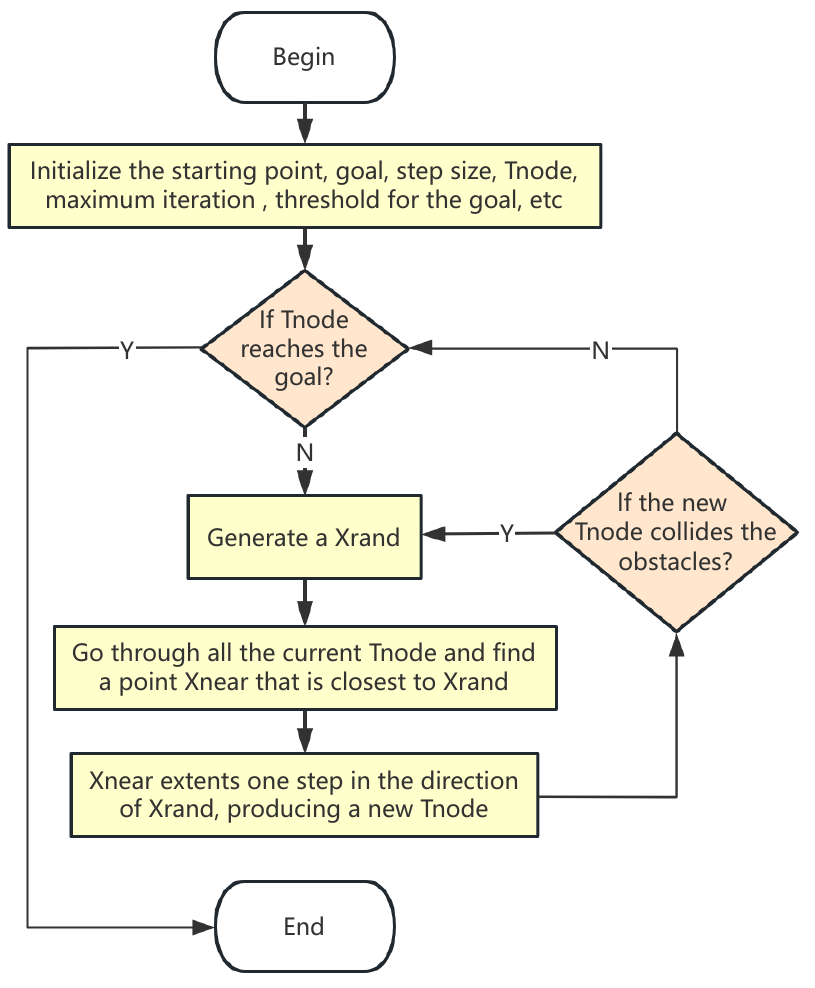}
    \caption{Workflow of RRT}
    \label{lct1}
\end{figure}

Rapidly-exploring Random Tree (RRT), proposed by LaValle and Kuffner in 1998, is an efficient sampling-based path planning algorithm \cite{bib22}. Initially developed for high-dimensional path planning, RRT can generate feasible paths from a starting point to a target under complex constraints. The design of RRT addresses the computational complexity faced by traditional methods in high-dimensional and obstacle-rich spaces. Given its efficient exploration capabilities, RRT has been widely applied in robotic motion planning, autonomous driving, UAV navigation, and medical robotic arm navigation. The core concept of RRT is to gradually grow a tree-like structure in the state space through random sampling, dynamically expanding the tree nodes to cover the free space. As shown in Figure~\ref{fig1}, the root node of the RRT tree is the starting point. A random sample point $X_{rand}$ is generated with a uniform probability distribution, and the nearest node $X_{near}$ is identified. Extending $X_{new}$ towards $X_{goal}$ by one step yields a new node. If $X_{new}$ is within feasible space, it is added to the tree. When $X_{new}$ is sufficiently close to the goal or meets a termination criterion, a feasible path from the start to the goal is found, and the algorithm terminates. Figure~\ref{lct1} presents the workflow of RRT.\par

\begin{figure}[ht]
    \centering
    \includegraphics[width=0.6\textwidth]{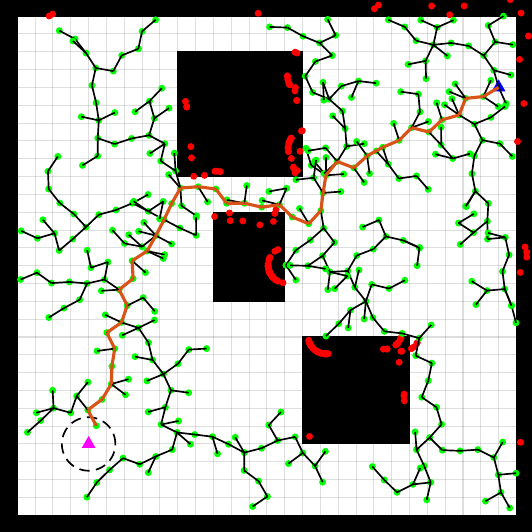}
    \caption{A path planned by the classic RRT in a 2D map}
    \label{fig3}
\end{figure}

The sampling strategy of RRT enables rapid, uniform exploration of unknown spaces, making it highly efficient in free space coverage. Figure~\ref{fig3} shows an RRT-planned path on a 2D map. RRT's performance in high-dimensional and obstacle-dense environments is exceptional, particularly suited for real-time applications. However, RRT has notable limitations:\par
\begin{itemize}
    \item \textbf{Path quality:} The generated paths are not optimal and tend to be winding.
    \item \textbf{Lack of convergence:} RRT does not guarantee an optimal solution and may fail to find the shortest path in complex environments.
    \item \textbf{Non-deterministic quality:} RRT's randomness can result in unstable path quality and, with limited samples, may not find a solution.
\end{itemize}
To address these limitations, this paper introduces an enhanced RRT (dRRT) to facilitate rapid path planning for UAVs in urban environments.

\subsection{dRRT}
\subsubsection{Target Bias Strategy}
The standard RRT algorithm generates new nodes through random sampling, expanding the search tree across space. However, this randomness, while fostering exploration diversity, can lead to inefficiencies, particularly in high-dimensional or complex environments where RRT may converge slowly due to local optima. Enhancing RRT's convergence efficiency and its search capacity toward the goal is a critical area for improvement.\par
The Target Bias Strategy addresses these shortcomings by introducing Target Bias Strategy, which, with a certain probability, selects the goal point as the sample point, thus increasing the likelihood that the search tree directly extends toward the target. In dRRT, a probability parameter $P_{target}$ controls the intensity of the target bias. Specifically, as shown in \eqref{eq1}, with a probability of $P_{target}$, the target point is directly set as the sample coordinate, thereby guiding the tree to expand toward the goal. If this probability is not met, a random sample is selected to maintain exploration diversity. The introduction of the target bias strategy largely overcomes RRT's tendency for lengthy paths and slow convergence due to purely random sampling. By creating a guiding effect toward the target, the search tree extends more directly in the goal's direction, significantly reducing the time needed to find a feasible path. Furthermore, this strategy reduces unnecessary path extensions during the search process, preventing the tree from aimlessly wandering near the target, thus lowering computation costs and improving search efficiency. In practical applications, target bias increases search density near the target, enhancing local search efficiency. This strategy demonstrates significant advantages in convergence and computational efficiency, suitable for path planning tasks in dynamic or obstacle-laden environments.\par
\begin{align}
    X_{sample}=\begin{cases}X_{goal},if\mathrm{~}rand<P_{target}\\X_{rand},else\end{cases}
    \label{eq1}
\end{align}
Where $X_{goal}$ is the target point coordinate; $T_{target}$ is the goal bias probability; $rand$ is a random number between [0, 1]; $X_{rand}$ represents a randomly generated point.

\subsubsection{Dynamic Step Size Strategy}
\begin{figure}[htbp]
    \centering
    \includegraphics[width=0.6\textwidth]{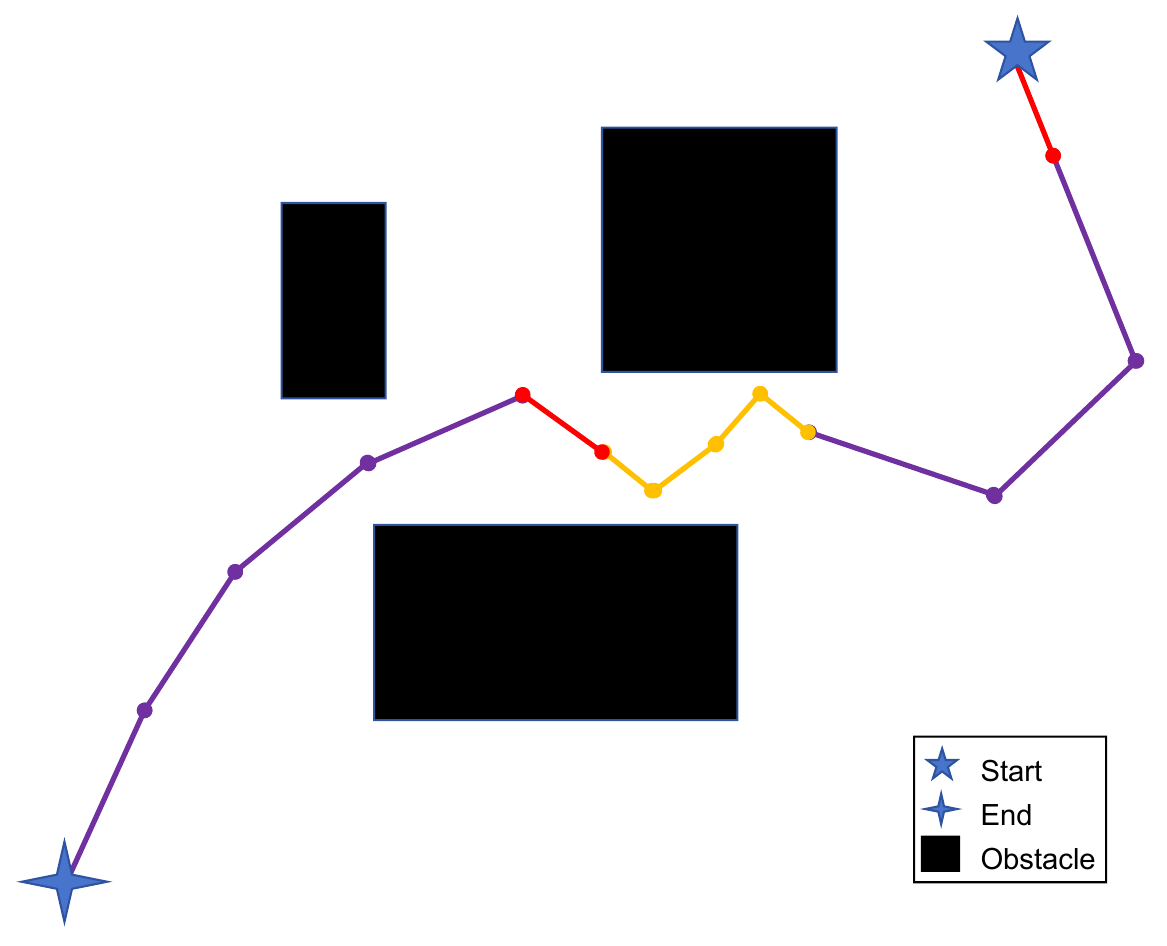}
    \caption{Simulation of Dynamic Step Size Strategy}
    \label{fig4}
\end{figure}

While the target bias strategy significantly enhances the search efficiency of the tree, it presents challenges in terms of fine detail path handling. When encountering obstacles in complex environments, a direct path toward the goal may be infeasible. Therefore, we introduce the Dynamic Step Size Strategy as shown in \eqref{eq2}. This strategy allows for larger step sizes in open spaces for rapid exploration, while reducing step size near obstacles to finely adjust the path direction, increasing the likelihood of navigating around obstacles. Dynamic Step Size Strategy can adjust the step size to explore the space around obstacles more flexibly, so that RRT can expand the path in different areas more carefully, and improve the feasibility of the path. Figure~\ref{fig4} is a simulation of the Dynamic Step Size Strategy.\par

\begin{equation}
    \resizebox{0.8\columnwidth}{!}{
    $StepSize2 = 
    \begin{cases}
        \min(StepSize_{max}, e \cdot StepSize2), & \text{if}\, X_{new} \text{ is away from obstacles}, \\
        \max(StepSize_{min}, StepSize2/e), & \text{if}\, X_{new} \text{ collides with obstacles}, \\
        StepSize, & \text{else}.
    \end{cases}$
    }
    \label{eq2}
\end{equation}

\subsection{Detour priority Strategy}
\begin{figure}[htbp]
    \centering
    \includegraphics[width=0.4\textwidth]{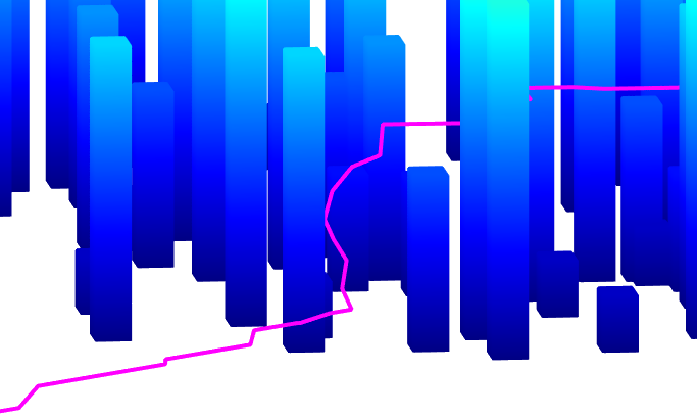}
    \caption{Vertical Ascension Phenomenon in RRT}
    \label{fig5}
\end{figure}

The classical RRT typically considers only the shortest straight-line connection for path extension. However, in real environments, due to the presence of obstacles, a direct path toward the target is often infeasible. The classical RRT lacks a mechanism for detouring around densely obstructed areas, which results in a phenomenon where the RRT often plans paths that vertically ascend rather than prioritizing obstacle avoidance, as illustrated in Figure~\ref{fig5}. This tendency increases the number of failed attempts and computational cost. Therefore, improving the classical RRT algorithm to guide it toward detour paths in densely obstructed areas is essential.\par
\begin{figure}[htbp]
    \centering
    \includegraphics[width=0.6\textwidth]{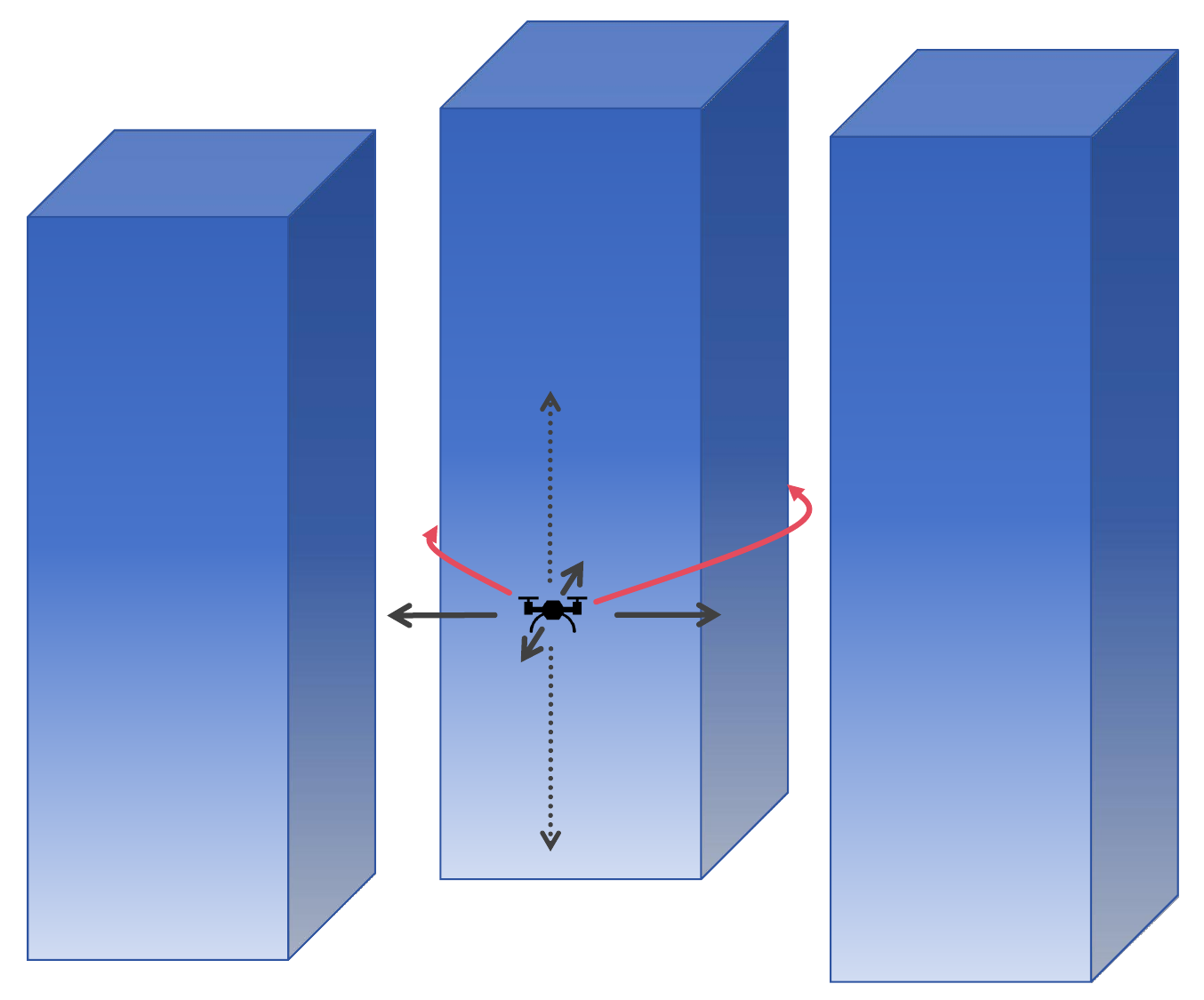}
    \caption{Simulation of Detour Priority Strategy}
    \label{fig6}
\end{figure}

To address this, we introduce a Detour Priority Strategy in dRRT, shown in Figure~\ref{fig6}. In this approach, when the tree's expansion direction is blocked by obstacles, the algorithm no longer forces an expansion towards the target but instead prioritizes finding detour paths in the XY plane. Specifically, four primary directions (forward, backward, left, and right) are defined, and feasibility checks are conducted for each direction to assess potential detour paths. The closest, obstacle-free direction is selected to incrementally bypass obstacles, thereby preventing the algorithm from becoming stuck. If none of these directions are feasible, the algorithm then decides whether the tree should expand upward or downward based on the current height and the target height. This strategy enhances dRRT's robustness in densely obstructed environments and significantly reduces instances of vertical ascension. By guiding the algorithm to prioritize detour paths, the Detour Priority Strategy substantially decreases the number of failed attempts and improves search efficiency. Additionally, this strategy enables dRRT to generate more flexible paths in complex environments, thus enhancing path planning success rates.\par
Detour Priority Strategy brings significant improvements to dRRT algorithm, particularly in terms of continuity and smoothness in path planning. By selecting appropriate detour strategies in densely obstructed areas, the algorithm can extend paths more smoothly and find optimal routes without increasing search costs.\par

\subsubsection{B-spline smoothing}
\begin{figure}[htbp]
    \centering
    \includegraphics[width=0.8\textwidth]{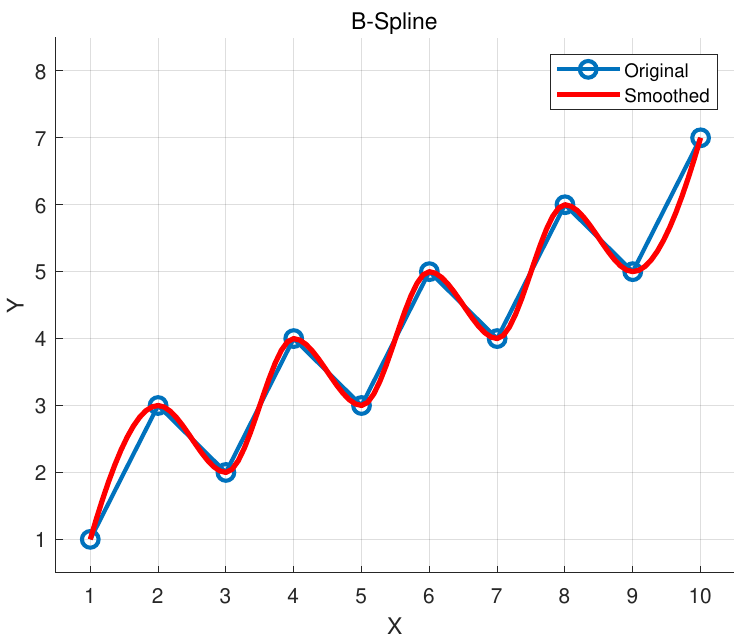}
    \caption{Comparison of the path before and after smoothing with B-spline}
    \label{fig8}
\end{figure}

Paths generated by the classical RRT algorithm typically consist of a series of discrete waypoints. While these waypoints form a feasible path from start to goal, the path is often irregular and unsmooth, making it unsuitable for practical applications. To enhance the path's smoothness and practicality, dRRT incorporates a B-spline smoothing strategy after path generation to further optimize the path. Figure~\ref{fig8} illustrates the effect of B-spline smoothing on a winding path; the blue line represents the original winding path, and the red line shows the smoothed path produced by B-spline smoothing. It is clear that after B-spline smoothing, the path's winding is substantially reduced, making the planned path better aligned with the robot's dynamics. \eqref{eq3} presents the formula for B-splines \cite{b11}.\par
\begin{equation}
    C(u)=\sum_{i=0}^nN_{i,k}(u)\cdot P_i
    \label{eq3}
\end{equation}
Where $C(u)$ is the coordinate of the curve at parameter $u$; $N_{i,k}(u)$ is the B-spline basis function of order $k$, with this paper using cubic B-splines, thus $k$=3; $P_i$ represents the $i^{th}$ control point; and $u$ is a parameter that usually varies within [0, 1].\par
\begin{figure*}[htbp]
    \includegraphics[width=\textwidth]{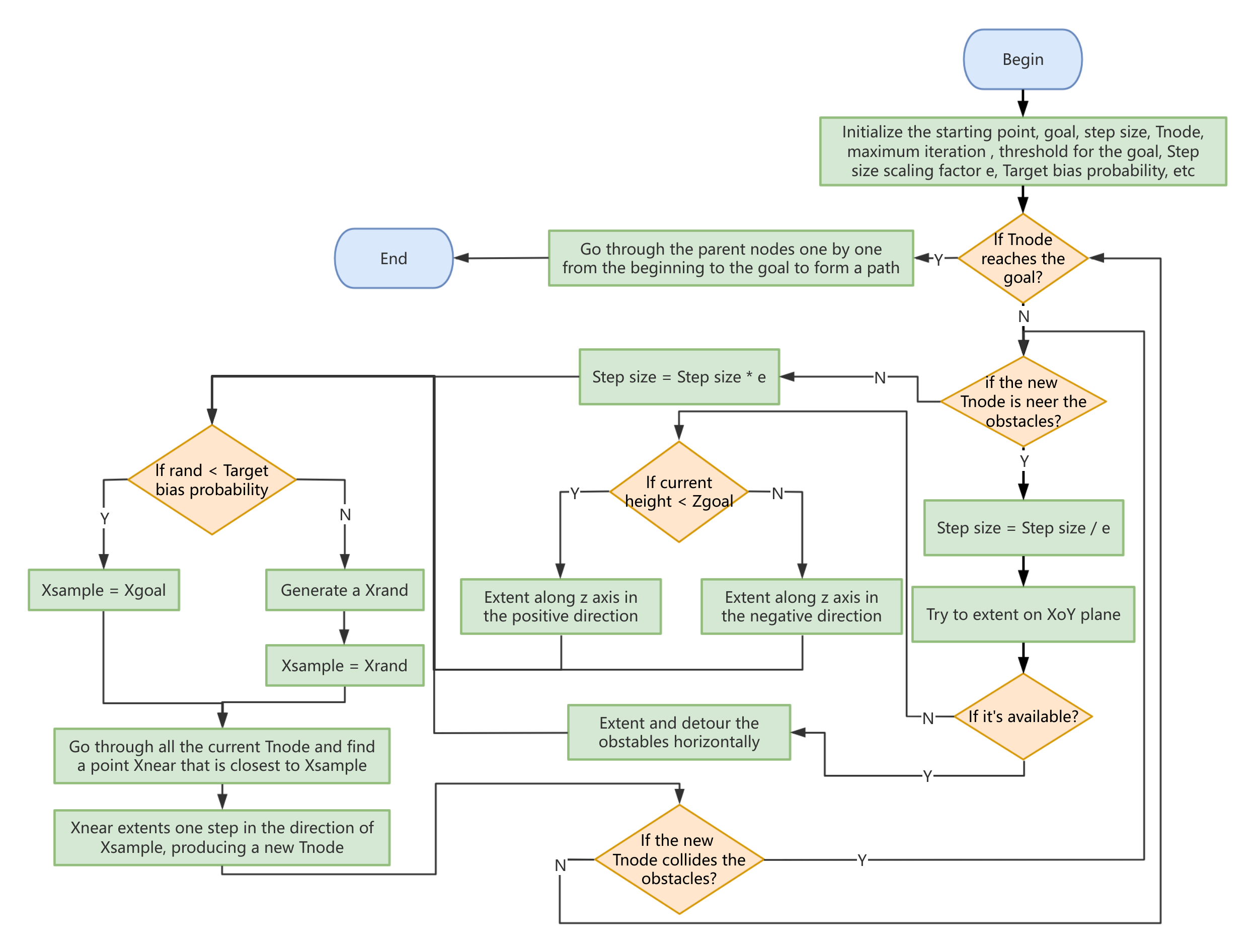}
    \caption{Workflow of dRRT}
    \label{lct2}
\end{figure*}

The fundamental concept behind B-spline smoothing is to construct a B-spline curve that fits the discrete nodes into a continuous, smoother curve, making the path more suitable for actual operations and motion control. B-spline curves exhibit excellent smoothness and adjustability, allowing path shape modifications by adjusting control points. In dRRT, a set of control nodes is extracted after path generation, and B-spline interpolation is applied to create a smoothed path, enhancing path continuity and adaptability to UAV dynamics. The smoothing process not only reduces sharp turns and unnecessary path nodes but also improves the path's operational viability. Incorporating B-spline smoothing into the dRRT algorithm provides a more refined and optimized solution for path planning, significantly increasing the practical applicability of the path. By integrating the B-spline smoothing strategy, dRRT paths are more feasible for practical operations, making them suitable for path planning tasks in various complex environments. Figure~\ref{lct2} is the workflow of dRRT.\par

\section{Simulation Experiments}
This section evaluates the performance of the dRRT through simulation experiments. To comprehensively validate dRRT's effectiveness and advantages in urban path planning, we compare it with classical path planning algorithms. First, we compare it with the traditional RRT algorithm to demonstrate the effectiveness of the proposed improvements; we then further compare dRRT with A* and Ant Colony Optimization (ACO) algorithms to verify dRRT's superior performance.\par
\subsection{Modeling of urban buildings}
\begin{figure}[htbp]
    \centering
    \includegraphics[width=0.6\textwidth]{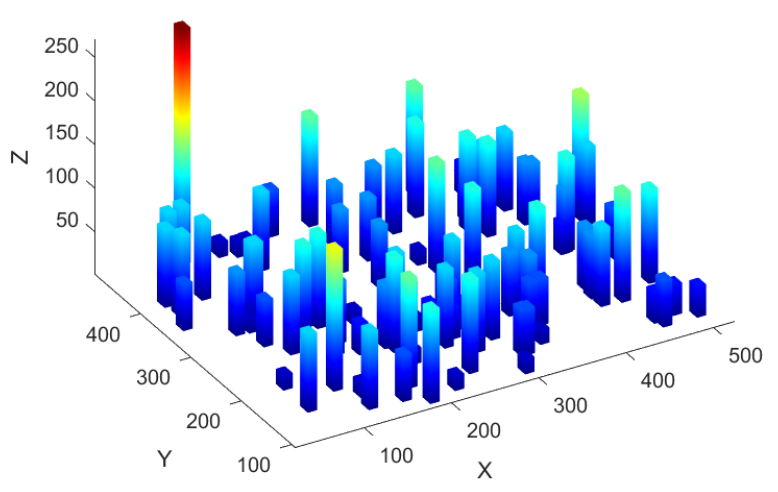}
    \caption{3D Urban Map Model}
    \label{fig9}
\end{figure}

In this paper, as illustrated in Figure~\ref{fig9}, we set the map range to (500m*500m*500m) and simulate the distribution of urban buildings within this area. Rectangular prisms represent buildings of various heights, ranging from 18m to 270m. The start coordinates are set to (10, 10, 1), and the target coordinates are set to (470, 420, 50).\par

\subsection{Comparison Between dRRT and RRT}
In urban path planning, the traditional RRT algorithm can generate a feasible path relatively quickly. However, the generated path often includes significant turns, is overly long, and requires a longer planning time, impacting the practical feasibility and efficiency of UAV flight paths. The proposed dRRT algorithm aims to improve path quality, speed up planning, reduce large turns along the path, and optimize computational efficiency. This experiment aims to demonstrate dRRT's advantages in the following aspects:\par
\begin{itemize}
    \item \textbf{Improved Path Quality:} Reducing path length and the occurrence of large turns, making the generated path smoother and more feasible.
    \item \textbf{Optimized Computational Efficiency:} Improving algorithm efficiency by reducing computation time and search space through search process enhancements.
    \item \textbf{Enhanced Operability:} Smoothing the path, particularly by controlling turns with yaw angles greater than 45°. Figure 10 illustrates the yaw angle; if the yaw angle is less than 45 degrees, the path turn is smoother, making the generated path easier to execute.
\end{itemize}

\begin{figure}[htbp]
    \centering
    \includegraphics[width=0.6\textwidth]{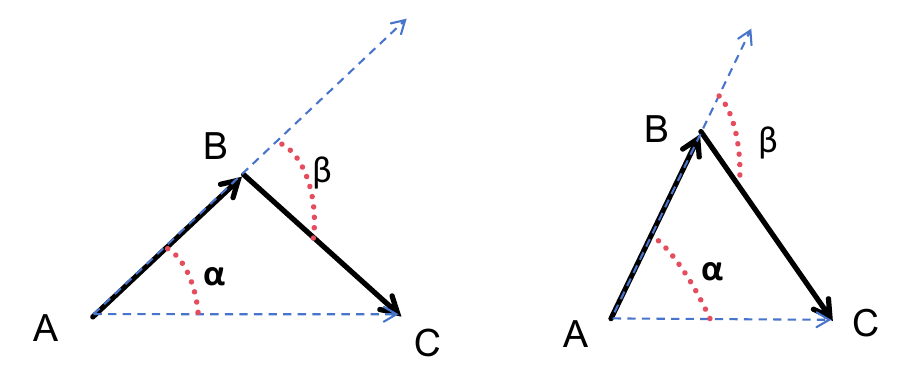}
    \caption{Yaw angle $\alpha$ and turning angle $\beta$}
    \label{fig10}
\end{figure}

\begin{figure}[htbp]
    \centering
    \includegraphics[width=0.8\textwidth]{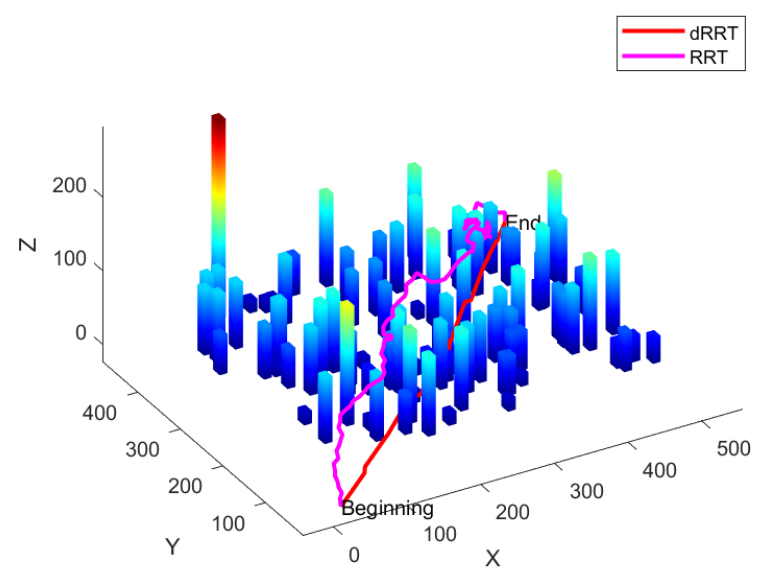}
    \caption{dRRT vs RRT}
    \label{fig11}
\end{figure}

Each algorithm was run 30 times in an urban map to eliminate randomness and verify robustness. The parameter settings for each algorithm are shown in Table~\ref{table1}. For each run, we recorded: time $t$, length $l$, smoothed length $l'$, number of waypoints $w$, total number of search grids $m$, search reward rate $\eta$, maximum yaw angle $\beta$, maximum yaw angles $\beta'$ after smoothing, frequency n of turning angles $\alpha$ exceeding 45°, and frequency $n'$ of turning angle $\alpha$ exceeding 45°after smoothing. The results are presented in Figure~\ref{fig11} and Table~\ref{table2}. Figure~\ref{fig10} is a description on  yaw angle $\alpha$ and turning angle $\beta$ 

\begin{table}[htbp]
    \centering
    \caption{Parameter setting}
    \begin{tabular}{|c|c|c|}
    \hline
    \textbf{Parameter} & \textbf{RRT} & \textbf{dRRT} \\
    \hline
    $StepSize$ & 10 & 10 \\
    \hline
    Threshold for reaching the goal & 5 & 5 \\
    \hline
    Max failed attempt & 20000 & 20000 \\
    \hline
    $P_{target}$ & - & 0.9 \\
    \hline
    Step size scaling factor \(e\) & - & 1.2 \\
    \hline
    $StepSize_{max}$ & - & 15 \\
    \hline
    $StepSize_{min}$ & - & 1 \\
    \hline
    \end{tabular}
    \label{table1}
\end{table}

\begin{table}[htbp]
    \centering
    \caption{The path planning results of dRRT and RRT}
    \begin{tabular}{|c|c|c|}
    \hline
    \textbf{Metrics} & \textbf{RRT} & \textbf{dRRT} \\
    \hline
    $t$(s) & 0.07971 & 0.01654 \\
    \hline
    $l$(m) & 997.3509 & 670.6704 \\
    \hline
    $l'$(m) & - & 670.6263 \\
    \hline
    $w$ & 104.16 & 94.36 \\
    \hline
    $m$ & 13199.23 & 483.88 \\
    \hline
    $\eta$ & 98.02\% & 100\% \\
    \hline
    $\beta$($^\circ$) & 97.9821 & 100.1452 \\
    \hline
    $\beta'$($^\circ$) & - & 91.8772 \\
    \hline
    $n$(times) & 50.68 & 11.64 \\
    \hline
    $n'$(times) & - & 7.6 \\
    \hline
    \end{tabular}
    \label{table2}
\end{table}

\subsection{Comparison of dRRT with A* and ACO Algorithm}
\begin{figure}[htbp]
    \centering
    \includegraphics[width=0.8\textwidth]{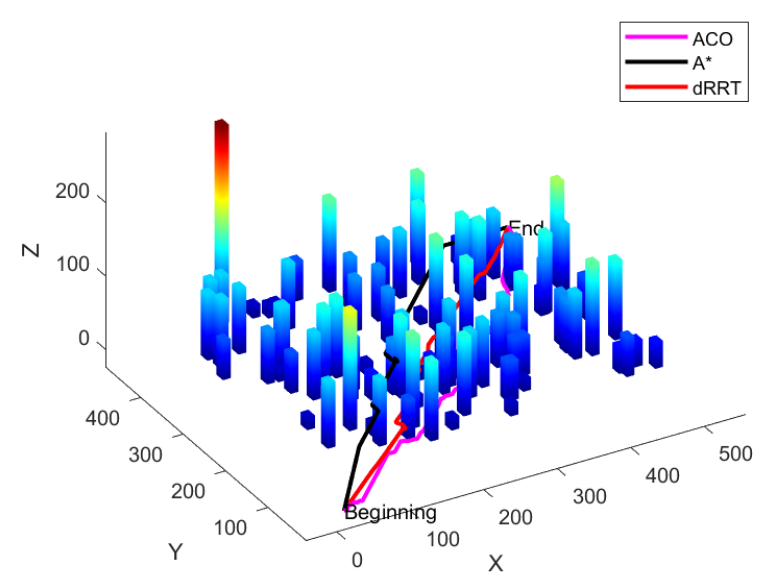}
    \caption{dRRT vs A* vs ACO}
    \label{fig12}
\end{figure}

A* and Ant Colony Optimization (ACO) are classical path planning algorithms widely recognized for their robust global optimization capabilities. A* is especially valued for its efficiency in global optimization and is broadly utilized in path search applications. ACO, on the other hand, models the foraging behavior of ants to accomplish global optimization, often applied for finding optimal paths. In this study, we aim to demonstrate that dRRT can offer faster computation speed and superior path quality under certain scenarios by comparing its performance against A* and ACO. Each algorithm was run 30 times in an urban map to eliminate randomness and verify robustness. For each run, we recorded: time $t$, length $l$, smoothed length $l'$, number of waypoints $w$, total number of search grids $m$, search reward rate $\eta$, maximum yaw angle $\beta$, maximum yaw angles $\beta'$ after smoothing, frequency n of turning angles $\alpha$ exceeding 45°, and frequency $n'$ of turning angle $\alpha$ exceeding 45°after smoothing. The results are presented in Fig.~\ref{fig12} and Table~\ref{table3}.\par
The results reveal that dRRT has an average runtime of 0.01468s, significantly outperforming both A* and ACO, indicating high computational efficiency. Among the three, dRRT generated the shortest paths. Furthermore, the number of waypoints produced by dRRT is substantially lower than those produced by A* and ACO, demonstrating a more streamlined and smoother path. The dRRT also explored fewer grid cells with average 887.6, highlighting a more efficient search space utilization. dRRT achieved a success rate of 100\%, outperforming both A* and ACO and showcasing strong robustness. The maximum yaw angle achieved by dRRT was the smallest among the three algorithms, and it had fewer instances of turning angles exceeding 45°, indicating the smoother path. After smoothing, sharp turns were further reduced, and the shortest path was achieved, indicating high-quality path planning by dRRT. Overall, in urban environments, dRRT demonstrated superior computational speed and path quality, particularly excelling over A* and ACO in computational efficiency and path smoothness. dRRT shows strong competitiveness for path planning applications, particularly in scenarios with high obstacle density and requirements for path smoothness, where it displays advantageous adaptability and robustness.

\begin{table}[htbp]
\caption{The path planning results of ACO, A* and dRRT}
\centering
\begin{tabular}{|c|c|c|c|}
\hline
\textbf{Metrics} & \textbf{ACO} & \textbf{A*} & \textbf{dRRT} \\
\hline
$t$ (s)& 18.417 & 0.2643 & 0.01468 \\
\hline
$l$(m) & 738.0945 & 667.6615 & 659.5079 \\
\hline
$l'$(m) & - & - & 659.2376 \\
\hline
$w$ & 538.33 & 550.67 & 93.87 \\
\hline
$m$ & 272511 & 14300 & 887.6 \\
\hline
$\eta$ & 64.18\% & 90\% & 100\% \\
\hline
$\beta$($^\circ$) & 109.4712 & 173.6598 & 102.7654 \\
\hline
$\beta'$($^\circ$) & - & - & 92.3524 \\
\hline
$n$(times) & 91 & 49 & 11.23 \\
\hline
$n'$(times) & - & - & 7.7 \\
\hline
\end{tabular}
\label{table3}
\end{table}

\section{Analysis}
This paper proposed an enhanced Rapidly-exploring Random Tree (dRRT) method for Unmanned Aerial Vehicle (UAV) path planning, which overcomes the limitations of traditional RRT algorithms in complex environments, achieving significant improvements in path convergence speed, feasibility, and smoothness. The dRRT introduces a Target Bias Strategy, Dynamic Step Size Strategy, Detour Priority Strategy, and B-spline smoothing strategy.\par
Target Bias Strategy accelerates convergence by increasing the probability of sampling points oriented towards the target, reducing excessive path generation. Dynamic Step Size Strategy enables automatic adjustment of step sizes in open versus obstacle-dense regions, enhancing the algorithm's adaptability to diverse environments. Detour Priority Strategy optimizes path expansion in dense obstacle areas, significantly reducing the number of invalid path explorations, effectively lowering computational costs and improving robustness. Finally, B-spline smoothing strategy ensures the practicality and operability of generated paths by smoothing them, which is especially valuable for UAV path planning applications.\par
In summary, the dRRT proposed in this study enhances path planning performance through the integration of multiple strategies, providing novel insights and technical support for efficient autonomous UAV navigation in complex environments.

\section{Limitations}
Although the dRRT algorithm has made significant improvements in path convergence speed, feasibility, and smoothness, it still has certain limitations and drawbacks. While dRRT optimizes the search strategy, the additional mechanisms such as goal biasing, dynamic step size, and detour priority introduce extra computational overhead, which can impact the performance in real-time applications with high timing requirements. The goal biasing strategy may cause the sampling points to be overly concentrated in the direction of the target, reducing the diversity of the search space and potentially causing the algorithm to fall into a local optimum, making it difficult to find the global optimum path in complex environments. Furthermore, dRRT requires integration with the Dynamic Window Approach (DWA) to achieve real-time obstacle avoidance.

\bibliographystyle{unsrt}  
\bibliography{references}

\end{document}